\documentclass[letterpaper, 10 pt, conference]{ieeeconf}  

\IEEEoverridecommandlockouts                              

\usepackage{graphics} 
\usepackage{graphicx}
\usepackage[backend=biber,style=ieee,natbib=true]{biblatex}
\usepackage{url}
\bibliography{bibliography}

\usepackage[font=small]{caption}

\title{Impact of Gaze-Based Interaction and Augmentation on \\ Human-Robot Collaboration in Critical Tasks}

\author{Ayesha Jena \and Stefan Reitmann \and Elin Anna Topp
\thanks{All authors are with the Department of Computer Science, Faculty of Engineering, Lund University, Box 118, S-221 00 Lund, Sweden, e-mail: \{ayesha.jena,stefan.reitmann,elin\_a.topp\}@cs.lth.se}%
}

\begin{document}

\maketitle
\thispagestyle{empty}
\pagestyle{empty}

\begin{abstract}
We present a user study analyzing head-gaze-based robot control and foveated visual augmentation in a simulated search-and-rescue task. Results show that foveated augmentation significantly improves task performance, reduces cognitive load by 38\%, and shortens task time by over 60\%. Head-gaze patterns analysed over both the entire task duration and shorter time segments show that near and far attention capture is essential to better understand user intention in critical scenarios. Our findings highlight the potential of foveation as an augmentation technique and the need to further study gaze measures to leverage them during critical tasks.
\end{abstract}



\section{INTRODUCTION}
\vspace*{-0.25em}

Advancements in the field of robotics have led to a need for seamless collaboration and effective communication between humans and robots in different scenarios~\cite{hentout2019human}. While communication methods such as speech and gestures have been extensively studied, they require explicit commands which limits their effectiveness in dynamic real-world interaction scenarios. In such cases, methods like gaze-based interaction offer an intuitive way of communication~\cite{lavit2024gaze}. In addition to being a method of interaction, gaze also shows operator's intentions regarding where to focus during task execution~\cite{belcamino2024gaze}.
This is crucial in high-stakes scenarios, where fast recognition of user intent through gaze could enhance performance and improve collaboration.

In human-robot collaboration, gaze tracking has proved effective in combination with augmentation techniques to improve collaboration efficiency, situational awareness, and productivity~\cite{schirmer2025utilizing}. While gaze is used for controlling the robot, augmentation techniques are used for visually enhancing the information provided by the system. Their effect in critical domains have been largely unexplored~\cite{sheridan2016human}, which leads to an adoption gap in understanding operator intentions and supporting collaboration in high-stake scenarios.





\begin{figure}
    \centering
    \includegraphics[width=0.8\linewidth]{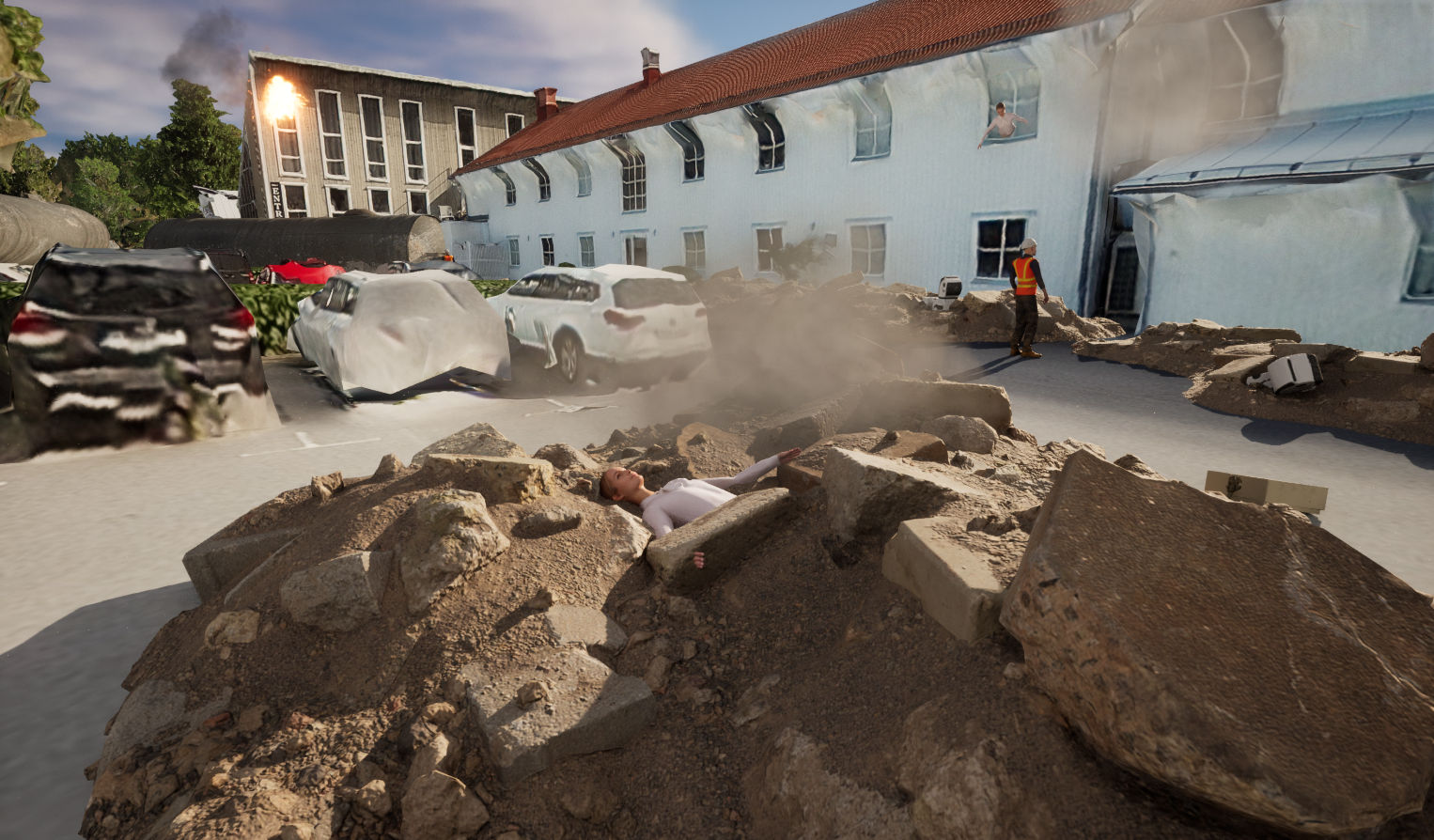}
    \caption{A region of the test scenario in simulation, showing a trapped human.}
    \vspace{-10pt}
    \label{fig:overview}
\end{figure}

We explore this gap through a user study using our collaborative interface~\cite{jena2023chaos,jena2024towards}.
This work is novel in its design and integration of head-gaze based interaction and real-time foveation-based visual augmentation for collaboration in a critical search-and-rescue scenario, as shown in Fig. \ref{fig:overview}. We will explain how we interpret the terms augmentation and foveation, in the context of our study in sections \ref{sec:relatedwork} and \ref{sec:methodology}.  

Our study explored two different interaction designs within the interface: manual head-gaze based control interface - human assisted (HA), and a dynamic foveation based interface - system assisted (SA). We found that foveation based augmentation enhances task performance, reduces cognitive load, and improves collaboration in comparison with direct gaze-based control. While head-gaze directly indicates user attention, it performed suboptimally as a primary control modality in critical scenarios. Our analysis, however, shows the usability of foveated augmentation to guide attention also in such scenarios. We further found instances of extrafoveal attention capture which would be accounted for in future studies with an adaptive system that incorporates complex gaze behavior.

\vspace{-10pt}
\section{RELATED WORK}
\label{sec:relatedwork}
\vspace*{-0.25em}
When human and robots collaborate in teams for critical tasks and missions, the cognitive capabilities of operators tend to decline over time. This is because of the inherent nature of such scenarios where rapid actions need to be taken while receiving, processing and combining information from multiple sources at the same time
~\cite{mirbabaie2019reducing}. This often leads to errors which in turn can provoke catastrophic outcomes~\cite{liu2024affects}. 
Both explicit and implicit modes of communication have been studied to minimize the operator's effort while maintaining effective information exchange during task execution.
Early research focuses on explicit communication modalities such as speech, gestures, and haptic interfaces~\cite{urakami2023nonverbal}. While these methods are effective in structured environments, they often lead to additional cognitive demands on operators, particularly in dynamic, high-stakes scenarios~\cite{unhelkar2020decision}. This has led to interest in implicit communication cues, such as gaze, which enable more natural and intuitive human-robot coordination~\cite{lavit2024gaze}. 

\begin{figure*}[ht!]
    \centering
    \includegraphics[width=0.8\linewidth]{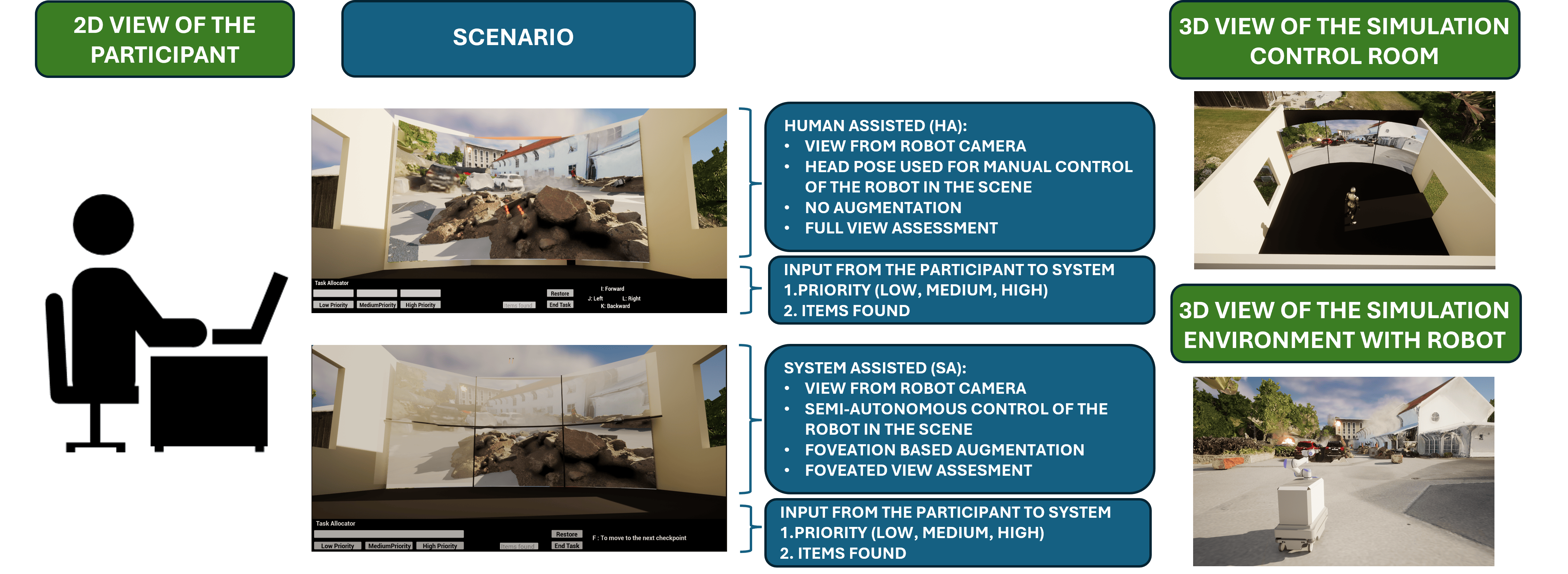}
    \caption{The setup of the user study. Left to right: 2D screen view of the interface shown to the participants during the experiment, 3D view of the simulation control room which receives visual input from the robot’s camera and provides inputs from the participants to the system, 3D view of the simulation environment showing the robot in the search-and-rescue test-bed.}
    \vspace{-10pt}
    \label{fig:virtual-environment}
\end{figure*}

Gaze-only interfaces encounter challenges like the ``Midas touch problem'', where unintentional gaze inputs trigger undesired actions, the need for special hardware and software for eye tracking, and difficulties in real-world scenarios due to illumination effects, head rotations, or occlusions~\cite{velichkovsky1997towards,jha2016analyzing}. 
A similar, yet effective, approach is to use head pose instead. Studies have shown a high correlation between gaze direction and head pose in real-world scenarios, proving its effectiveness~\cite{jha2016analyzing}.

Building on methods to improve collaboration through the use of a visual interface, we explored different augmentation techniques that would enhance task performance while reducing cognitive overload~\cite{su2023integrating}. Techniques like highlighting important objects seemed ineffective in our case due to clutter and background distractions. Instead, foveation seemed to be a promising approach in this regard where focused regions are rendered in high resolution and outlying areas are blurred~\cite{su2023integrating}. Traditionally, this is used as a graphics-performance optimization technique, so this study is the first to apply foveation as a visual augmentation method in a search-and-rescue setting.
Considering the complexities of real-world situations, this study investigates head-gaze based control and foveation based augmentation and analyzes the results using task performance metrics, subjective measurements, and gaze-based heatmaps.

\vspace{-2pt}
\section{Methodology}
\label{sec:methodology}
\vspace*{-0.25em}

The study was conducted in a simulation environment where participants performed a search task with two interaction designs: HA and SA. 
Participants’ head-gaze behavior was recorded and analyzed to assess efficiency, cognitive workload, and task performance. Head-gaze tracking was performed using a camera-based tracking system that detected participants’ head orientation and head-gaze direction. 
An overview and breakdown of the experimental conditions are shown in Fig. \ref{fig:virtual-environment}.

\vspace{-3pt}
\subsection{Simulation Environment}
A search-and-rescue (SAR) test environment ($28$ X $83$ m) was developed from the 3D template of the real-world location provided by the WASP Research Arena for Public Safety (WARA-PS)~\cite{cesium} to simulate a post-disaster scenario using Unreal Engine 5.1~\cite{unrealengine}. The primary motivation behind creating this virtual SAR test-bed was to replicate challenging and high-stake conditions in a controlled, safe environment. The added wreckage, obstacles, and people were strategically placed throughout the environment. It also had multiple other areas of interest, such as fire outbreaks and electrical hazards. Additionally, it included simulated SAR personnel to replicate realistic operational constraints and potential coordination efforts in real-time rescue missions. This was done on a collaborative interface developed earlier~\cite{jena2023chaos,jena2024towards}. The interface sends gaze data to Unreal Engine, where it’s mapped with 3D vectors to highlight screen regions in orange. A key press turns the region green, forming a dual-confirmation input. It also includes text fields and buttons for selecting priorities, and overlays foveation cues on the camera feed from the robot (Fig. \ref{fig:virtual-environment}). Some distance away from the test area, a control room, as shown in Fig. \ref{fig:virtual-environment}, was designed to replicate real-world SAR operation centers where users could interact with the disaster site by navigating a mobile robot fitted with cameras and sensors.
\vspace{-5pt}
\subsection{Experiment Procedure}
Each session lasted about 60 minutes and began with a system check, consent and demographic forms, and a brief training. Participants then completed two counterbalanced scenarios - HA and SA, each involving teleoperating the robot through the interface. In each scenario, they navigated within the test environment to find areas that needed inspection also referred to as areas of interests (AOIs), counted and classified objects (victims, debris, hazards), and assigned priority levels (Low, Medium, High) through the interface as can be seen in Fig. \ref{fig:virtual-environment}. Before the experiment it was made sure that the ethical protocols were followed per university guidelines~\cite{luEthicalReview}. During each session, we recorded head-gaze, dual-confirmation key presses, robot state transformations, NASA-TLX~\cite{hart1988development}, System Usability Scale (SUS)~\cite{brooke1996sus}, post-study interview responses, and quantitative metrics (task times, AOI accuracy), all anonymized under data protection rules.

A total of 18 participants (13 males, 4 females, 1 unspecified) took part in the study. 4 participants had experience in providing disaster
relief. The mean age of participants was 31.29 years (SD = 9.78) excluding 1 participant who declined to report their age. 

\subsubsection{Human-Assisted (HA) Scenario}
In this interaction design scenario, participants controlled the robot using head-gaze for directions and keyboard confirmations.

\subsubsection{System-Assisted (SA) Scenario}

\begin{figure*}[htp]
    \centering
    \includegraphics[width=0.8\linewidth]{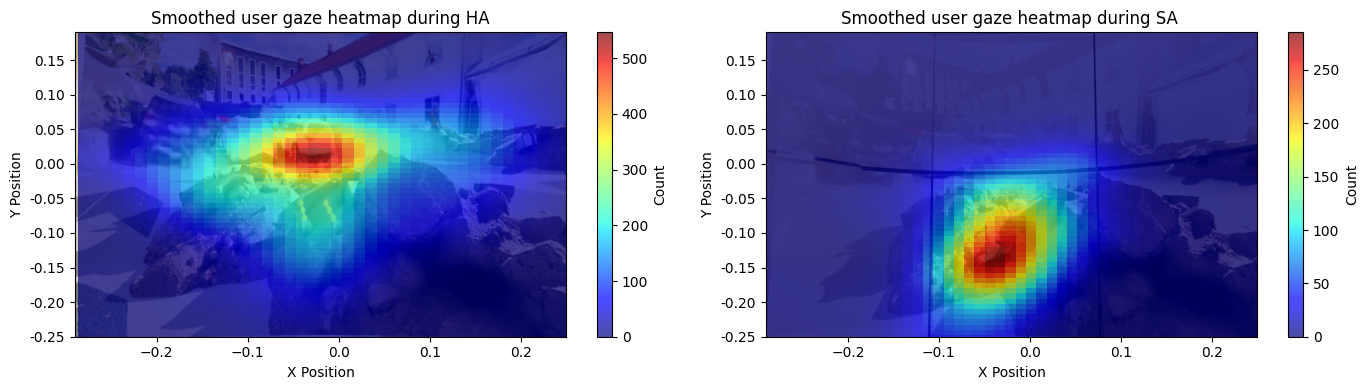}
    \caption{Distribution of head-gaze heatmap over the 2D screen area presented to the participants.}
    \label{fig:heatmap}
\end{figure*}

\begin{table*}[h!]
    \caption{Objective and Subjective Results for HA and SA scenarios. $*$ indicates significant result ($p<0.001$)}
    \centering
    \begin{tabular}{|c|ccccc|}
    \hline
         Scenario & Total Time Taken (in s) & Total Humans Saved & Avg Humans Saved & NASA TLX Score & SUS\\ \hline
         HA & $678.88 \pm 233.98$ & $20$ (Out of 54)  & $1.11 \pm 0.75$ & $53.85 \pm 18.06$ & $58.61 \pm 14.80$ \\
         SA & $\ast\textbf{274.41} \pm \textbf{52.95}$ & $\ast\textbf{47}$ (Out of 54) & $\ast\textbf{2.61} \pm \textbf{0.69}$ & $\ast\textbf{33.4} \pm \textbf{15.24}$ & $\ast\textbf{80.13} \pm \textbf{16.30}$ \\
         \hline
    \end{tabular}
    \vspace{-10pt}
    \label{tab:sa_vs_ha_table}
\end{table*}

In this scenario, the 6 screens shown to the participants were foveated based on the AOIs present and the density of hazard. Screens with AOIs were in focus whereas others were blurred. Although the system guided participants to prioritized regions, participants remained the final decision-makers, confirming or adjusting the importance levels of each item in AOIs (Low, Medium, High) through the interface. The six-section design was optimized to align with human visual working memory limits (5–9 chunks~\cite{miller1956magical}), ensuring rapid parsing without overwhelming users during critical scenarios.



\vspace{-5pt}
\section{Results and Discussions}
\vspace*{-0.25em}
In both the scenarios, participants were instructed to identify critical AOIs, mark them, and assign points according to their priority. The results are discussed below.
\vspace{-5pt}
\subsection{Performance Metrics}
The performance of the participants was measured using task completion time and the number of humans successfully located within AOIs. There were 3 trapped humans in each scenario within AOIs, resulting in a total of 54 instances of trapped humans across all participants. From the results shown in Table \ref{tab:sa_vs_ha_table}, we can see that participants performed better with foveation. The experience of the 4 participants with disaster relief did not affect the results in any manner.

\vspace{-5pt}
\subsection{Subjective Measurements}
\subsubsection{Mental Workload}
The participants mental workload results (Table \ref{tab:sa_vs_ha_table}) from the NASA TLX questionnaire showed a $38 \%$ lower perceived workload in SA scenario ($M = 33.41$, $SD = 15.24$) compared to the HA scenario  ($M = 53.85$, $SD = 18.07$).

\subsubsection{System Usability} 
Participants answered 10 questions from the system usability questionnaire using a Likert scale (1 = strongly disagree to 5 = strongly agree). Results from the Table \ref{tab:sa_vs_ha_table} show that participants perceived foveation to have higher usability during the search task.

\subsubsection{Subjective Questionnaire}
The subjective questionnaires contained long-form answers and five-point Likert scale questions. Participants provided neutral-to-positive ratings for the interface’s natural intuitiveness, noting that its reliance on manual gaze-driven navigation mirrored ``natural human exploration". However, this familiarity with gaze-based control came with challenges such as ``slow turning" and ``limited peripheral vision". On the other hand in SA, augmentation helped them streamline focus and execute the task faster, while reducing their cognitive load. However, participants also expected explainability regarding the augmentation decisions made by the system. In addition, participants wanted to be able to dynamically look at other sections of the screen to make sure the system made the correct decisions regarding foveation.
\vspace{-5pt}
\subsection{Head-gaze Analysis}
In the HA scenario, participants' head-gaze was evenly spread across the screen, looking equally to the left and right, but with a stronger focus on the upper part. This scenario required them to use their head-gaze to drive the robot, causing them to scan the screen more and focus on the upper part for forward navigation based on the interface design. In the SA scenario, participants did not need to orient their head-gaze, so they adopted a more relaxed posture and looked mostly at the lower part of the screen as can be seen in Fig. \ref{fig:heatmap}. Additionally, analyzing shorter time segments in this scenario during task execution highlight that gazed region and the foveated region aligned 67\% of the time. 
In the remaining 33\%, participants’ gaze slightly deviated around the foveated area. Upon further analysis of view counts, view percentages, and average gaze position across the screen area, it was found that this occurred due to high-priority items and people located outside the immediate field of view. This observation can be explained by extrafoveal attention capture which is when objects or individuals in more distant areas draw the user’s attention over a longer distance~\cite{nuthmann2019extrafoveal}. Similar instances of extrafoveal attention captures were also observed in the HA scenario where users verbally inquired about denoting distant AOIs in comparison to focusing on immediate AOIs.
\vspace{-7pt}
\section{Conclusion}
\vspace*{-0.25em}
In summary, we conducted a user study to investigate the impact of gaze-based control and foveated augmentation for human-robot teleoperation during critical task execution. The study was performed with a system designed in simulation and the effects were studied through two scenarios. Based on the results of the user study, foveation based augmentation scenario outperformed gaze-based control leading to effective human-robot collaboration.
Additionally, analysis of head-gaze behavior revealed alignment between gaze and foveated regions in a majority of cases, indicating that foveation can effectively direct attention. However, deviations due to extrafoveal attention capture highlight the complexity of user attention patterns and suggest the need for systems to accommodate such behavior. 

Overall, this study demonstrates that head-gaze based foveated augmentation improves performance and user experience. 
While head-gaze reliably conveys where the user’s attention is directed, our results suggest it is not optimal as a direct control modality in critical tasks. Further analysis is required to leverage gaze measures during critical tasks. The future work includes improving the interface by combining automated scene analysis with flexible, gaze dependent foveation that operators can invoke as an explicit confirmation cue.



\vspace{-5pt}
\section{Acknowledgement}
We would like to thank Jacek Malec and Björn Olofsson for their support during the work. This work was funded by ELLIIT - the Excellence Center at Linköping University and Lund University for Information Technology, and partially supported by the Wallenberg AI, Autonomous Systems and Software Program (WASP) funded by the Knut and Alice Wallenberg Foundation. We would also like to thank the WASP Research Arena for Public Safety (WARA-PS) for providing the 3D template.
\vspace{-5pt}
\printbibliography

\end{document}